\documentclass[11pt]{article}

\usepackage[final]{acl}

\usepackage{times}
\usepackage{latexsym}

\usepackage{microtype}
\usepackage{float}
\usepackage{xurl}

\usepackage[T1]{fontenc}
\usepackage[utf8]{inputenc}
\usepackage{microtype}
\usepackage{graphicx}
\usepackage{amsmath}
\usepackage{amssymb}
\usepackage{booktabs}
\usepackage{multirow}
\usepackage{xcolor}
\usepackage{subcaption}
\usepackage{enumitem}
\setlist{nosep,leftmargin=*}
\usepackage{pifont}

\usepackage{tikz}
\usetikzlibrary{shapes.geometric, arrows.meta, positioning, fit, backgrounds, calc}

\newcommand{\sys}{\textsc{FinGround}}

\newcommand{\eg}{\textit{e.g.}}
\newcommand{\ci}[1]{\textsubscript{\scriptsize$\pm$#1}}

\title{FinGround: Detecting and Grounding Financial Hallucinations\\via Atomic Claim Verification}

\author{
	Dongxin Guo$^{1}$, Jikun Wu$^{2}$, Siu Ming Yiu$^{1}$ \\
	$^{1}$The University of Hong Kong \quad $^{2}$Stellaris AI Limited \\
	\texttt{bettyguo@connect.hku.hk}, \texttt{hk950014@connect.hku.hk}, \\
	\texttt{smyiu@cs.hku.hk}
}

\begin{document}
\maketitle

\begin{abstract}
	Financial AI systems must produce answers grounded in specific regulatory filings, yet current LLMs fabricate metrics, invent citations, and miscalculate derived quantities. These errors carry direct regulatory consequences as the EU AI Act's high-risk enforcement deadline approaches (August 2026). Existing hallucination detectors treat all claims uniformly, missing 43\% of computational errors that require arithmetic re-verification against structured tables. We present \sys{}, a three-stage verify-then-ground pipeline for financial document QA. Stage~1 performs finance-aware hybrid retrieval over text and tables. Stage~2 decomposes answers into atomic claims classified by a six-type financial taxonomy and verified with type-routed strategies including formula reconstruction. Stage~3 rewrites unsupported claims with paragraph- and table-cell-level citations. To cleanly isolate verification value from retrieval quality, we propose \textit{retrieval-equalized evaluation} as standard methodology for RAG verification research: when all systems receive identical retrieval, \sys{} still reduces hallucination rates by 68\% over the strongest baseline ($p < 0.01$). The full pipeline achieves a 78\% reduction relative to GPT-4o. An 8B distilled detector retains 91.4\% F1 at 18$\times$ lower per-claim latency, enabling \$0.003/query deployment, supported by qualitative signals from a four-week analyst pilot.
\end{abstract}
\section{Introduction}
\label{sec:intro}

Financial professionals require answers grounded in specific regulatory filings and earnings reports, yet LLMs routinely fabricate financial metrics, invent regulatory citations, and distort ratios. GPT-4-Turbo with retrieval incorrectly answered or refused 81\% of curated SEC filing questions \citep{islam2023financebench}, with systematic fabrication of financial metrics documented across models \citep{kang2023deficiency, gartner2024ai, accenture2024financial}. The EU AI Act \citep{euaiact2024} mandates compliance for high-risk financial AI systems by August 2026 \citep{dahl2024large}, requiring human oversight with interpretable outputs (Article~14) and accuracy guarantees (Article~15).

Prior work has advanced individual components: general hallucination detection \citep{manakul2023selfcheckgpt, farquhar2024detecting, min2023factscore, wei2024safe}, financial QA benchmarks \citep{chen2021finqa, zhu2021tatqa, islam2023financebench}, and RAG systems \citep{asai2024selfrag, yan2024corrective}. Yet no existing system unifies detection and mitigation into a production-ready pipeline for financial QA. The financial domain breaks assumptions of prior systems in specific, quantifiable ways. FActScore and SAFE decompose claims into atomic facts but treat all facts uniformly, so they cannot verify ``gross margin was 62.4\%'' against table cells without structured extraction; this gap accounts for 43\% of computational errors that a domain-specific pipeline catches (Appendix~\ref{app:gap_experiment}). RARR's regeneration assumes single-source evidence, but when applied without claim-type-aware routing, 34\% of computational claim regenerations produced \textit{new} hallucinations. And table-cell attribution without structure-aware upstream chunking produced 23\% dangling citations pointing to misaligned cells.

We present \sys{}, addressing this gap through three contributions:

\begin{enumerate}[leftmargin=1.2em]
\item \textbf{Finance-Aware Atomic Verification}: decomposition of LLM answers into atomic claims verified against evidence using a validated six-type taxonomy (\textit{numerical}, \textit{temporal}, \textit{entity-attribute}, \textit{comparative}, \textit{regulatory}, \textit{computational}) with type-routed verification strategies including arithmetic re-computation (\S\ref{sec:verification}).

\item \textbf{Grounded Regeneration with Evidence Attribution}: targeted rewriting of only hallucinated spans with paragraph- and table-cell-level citations, achieving 93.2\% faithfulness on regenerated claims (\S\ref{sec:regeneration}).

\item \textbf{Efficient Distilled Detector}: distillation from GPT-4o into an 8B model achieving 91.4\% F1 at 18$\times$ lower per-claim latency, enabling \$0.003/query deployment (\S\ref{sec:distillation}).
\end{enumerate}

\noindent Beyond system contributions, we introduce \textit{retrieval-equalized evaluation}, in which all baselines receive identical retrieval, isolating verification value from retrieval gains. Under these conditions, atomic verification yields 68--76\% additional HalRate reduction ($p < 0.01$). Cross-generator evaluation on Llama-3-70B and Claude-3.5-Sonnet shows 87--89\% F1 transfer, and a four-week pilot with 24 analysts provides deployment design signals targeting EU AI Act compliance.

\section{Related Work}
\label{sec:related}

Hallucination detection spans sampling-based methods \citep{manakul2023selfcheckgpt, farquhar2024detecting}, atomic evaluation \citep{min2023factscore, wei2024safe, wang2024factcheckbench}, and industrial classifiers \citep{patronus2024lynx, galileo2024chainpoll, microsoft2024groundedness}, with comprehensive taxonomies established by surveys \citep{ji2023survey, huang2025survey, tonmoy2024comprehensive}. These approaches operate domain-agnostically without financial-specific claim typing or table-cell attribution.

Financial QA demands multi-step numerical reasoning over hybrid text-and-table content, with benchmarks from FinQA \citep{chen2021finqa} and TAT-QA \citep{zhu2021tatqa} through FinanceBench \citep{islam2023financebench} and DocFinQA \citep{reddy2024docfinqa} establishing evaluation standards. RAG variants such as Self-RAG \citep{asai2024selfrag}, CRAG \citep{yan2024corrective}, and Adaptive-RAG \citep{jeong2024adaptive}, together with attribution methods \citep{gao2023rarr, gao2023alce, bohnet2022attributed}, provide grounding infrastructure, while domain models \citep{wu2023bloomberggpt, yang2023fingpt} reduce but do not eliminate hallucination. Financial hallucination benchmarks PHANTOM \citep{ji2025phantom} and FAITH \citep{zhang2025faith} document severity but do not offer integrated solutions. None of these systems combine atomic claim verification with financial table-cell attribution and hallucination-triggered regeneration.

Industrial financial AI platforms (Bloomberg, Kensho, AlphaSense, FactSet) excel at information retrieval and extraction but do not perform claim-level verification: they cannot determine whether a specific LLM-generated assertion is supported by source evidence. \sys{} operates as a verification layer downstream of any financial QA system. Knowledge distillation \citep{hinton2015distilling, gu2024minillm} and demonstrated success transferring hallucination detection to 8B scale \citep{su2024mind} enable the efficient deployment our system requires. A detailed capability comparison is in Appendix~\ref{app:comparison}.

\section{The \sys{} System}
\label{sec:system}

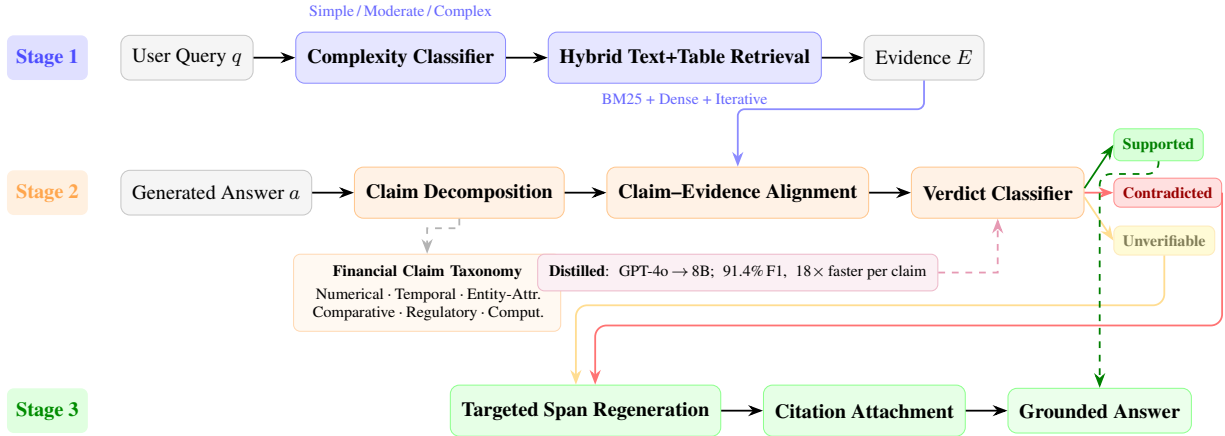
\begin{figure*}[t]
	\centering
	\resizebox{\textwidth}{!}{%
		\begin{tikzpicture}[
			node distance=0.55cm and 0.65cm,
			>=Stealth,
			every node/.style={font=\small},
			]
			
			\tikzset{
				stg/.style={rounded corners=4pt, minimum height=0.8cm, text centered,
					font=\small\bfseries, minimum width=1.8cm, inner sep=5pt},
				sA/.style={stg, fill=blue!10, draw=blue!45},
				sB/.style={stg, fill=orange!10, draw=orange!45},
				sC/.style={stg, fill=green!10, draw=green!45},
				iobox/.style={rounded corners=4pt, fill=gray!8, draw=gray!45,
					minimum height=0.7cm, font=\small, inner sep=5pt},
				myarr/.style={->, thick, >=Stealth},
				mydarr/.style={->, thick, dashed, gray!60, >=Stealth},
				tbox/.style={rounded corners=3pt, fill=orange!5, draw=orange!30,
					font=\scriptsize, inner sep=5pt},
				vbox/.style={rounded corners=3pt, font=\scriptsize, inner sep=4pt,
					minimum width=1.2cm, text centered},
				stglabel/.style={font=\footnotesize\bfseries, rounded corners=3pt,
					minimum width=1.0cm, inner sep=4pt, minimum height=0.7cm},
				annot/.style={font=\scriptsize, inner sep=2pt},
			}
			
			\node[stglabel, fill=blue!12, text=blue!70] (s1lab) {Stage 1};
			
			\node[iobox, right=0.5cm of s1lab] (query) {User Query $q$};
			\node[sA, right=0.65cm of query] (complex) {Complexity Classifier};
			\node[sA, right=0.65cm of complex, minimum width=2.8cm] (retrieval)
			{Hybrid Text{+}Table Retrieval};
			\node[iobox, right=0.65cm of retrieval] (evidence) {Evidence $E$};
			
			\draw[myarr] (query) -- (complex);
			\draw[myarr] (complex) -- (retrieval);
			\draw[myarr] (retrieval) -- (evidence);
			
			\node[annot, above=0.12cm of complex, color=blue!60]
			{Simple\,/\,Moderate\,/\,Complex};
			\node[annot, below=0.08cm of retrieval, color=blue!60]
			{BM25 + Dense + Iterative};
			
			\node[stglabel, fill=orange!12, text=orange!70,
			below=1.4cm of s1lab] (s2lab) {Stage 2};
			
			\node[iobox, right=0.5cm of s2lab] (answer) {Generated Answer $a$};
			\node[sB, right=0.65cm of answer] (decomp) {Claim Decomposition};
			\node[sB, right=0.65cm of decomp, minimum width=2.8cm] (align)
			{Claim--Evidence Alignment};
			\node[sB, right=0.65cm of align] (verdict) {Verdict Classifier};
			
			\draw[myarr] (answer) -- (decomp);
			\draw[myarr] (decomp) -- (align);
			\draw[myarr] (align) -- (verdict);
			
			\draw[myarr, blue!45, rounded corners=3pt]
			(evidence.south) -- ++(0,-0.45) -| (align.north);
			
			\node[vbox, fill=green!15, draw=green!45,
			right=0.45cm of verdict, yshift=0.75cm] (sup)
			{\textcolor{green!50!black}{\textbf{Supported}}};
			\node[vbox, fill=red!12, draw=red!45,
			right=0.45cm of verdict] (con)
			{\textcolor{red!60!black}{\textbf{Contradicted}}};
			\node[vbox, fill=yellow!20, draw=yellow!50!orange!30,
			right=0.45cm of verdict, yshift=-0.75cm] (unv)
			{\textcolor{yellow!40!black}{\textbf{Unverifiable}}};
			
			\draw[myarr, green!50!black]
			([yshift=2pt]verdict.east) -- (sup.west);
			\draw[myarr, red!55]
			(verdict.east) -- (con.west);
			\draw[myarr, yellow!50!orange!50]
			([yshift=-2pt]verdict.east) -- (unv.west);
			
			
			\node[tbox, text width=3.8cm, align=center,
			below=0.55cm of decomp, xshift=-0.5cm] (tax) {%
				\textbf{Financial Claim Taxonomy}\\[2pt]
				Numerical\,$\cdot$\,Temporal\,$\cdot$\,Entity-Attr.\\
				Comparative\,$\cdot$\,Regulatory\,$\cdot$\,Comput.};
			\draw[mydarr] (decomp.south) -- ++(0,-0.15) -| (tax.north);
			
			\node[rounded corners=3pt, fill=purple!6, draw=purple!28,
			font=\scriptsize, inner sep=5pt,
			below=0.55cm of align] (distill) {%
				\textbf{Distilled}:\; GPT-4o\,$\to$\,8B;\;
				91.4\%\,F1,\; 18$\times$\,faster per claim};
			\draw[mydarr, purple!40] (distill.east) -| (verdict.south);
			
			\node[stglabel, fill=green!12, text=green!55!black,
			below=2.7cm of s2lab] (s3lab) {Stage 3};
			
			\coordinate (mid23) at ($(decomp.east)!0.5!(align.west)$);
			\node[sC, minimum width=2.8cm] at (mid23 |- s3lab) (regen)
			{Targeted Span Regeneration};
			\node[sC, right=0.65cm of regen, minimum width=2.2cm] (cite) 
			{Citation Attachment};
			\node[iobox, right=0.65cm of cite, fill=green!8, draw=green!45,
			font=\small\bfseries] (final) {Grounded Answer};
			
			\draw[myarr] (regen) -- (cite);
			\draw[myarr] (cite) -- (final);
			
			\draw[myarr, red!55, rounded corners=5pt]
			(con.east) -- ++(0,-2.08) -| ([xshift=0.15cm]regen.north);
			
			\draw[myarr, yellow!50!orange!50, rounded corners=5pt]
			(unv.south) -- ++(0,-0.78) -| ([xshift=-0.15cm]regen.north);
			
			\draw[myarr, green!50!black, dashed, rounded corners=5pt]
			(sup.south) -- ++(0,-0.2) -| (final.north);
			
		\end{tikzpicture}%
	}
	\caption{The \sys{} pipeline. \textbf{Stage~1} classifies query complexity and retrieves hybrid text-and-table evidence. \textbf{Stage~2} decomposes answers into atomic financial claims, aligns each to evidence, and classifies verdicts using a distilled 8B model. \textbf{Stage~3} rewrites contradicted/unverifiable claims with cited evidence; supported claims pass through unchanged.}
	\label{fig:architecture}
\end{figure*}

\sys{} operates as a three-stage pipeline (Figure~\ref{fig:architecture}): (1)~finance-aware retrieval extracts evidence from hybrid text-and-table documents; (2)~atomic verification decomposes answers into claims and detects hallucinations; (3)~grounded regeneration rewrites unsupported claims with cited evidence.

\subsection{Stage 1: Finance-Aware Hybrid Retrieval}
\label{sec:retrieval}

Financial answers frequently require synthesizing information from narrative text and structured tables. Adapting query-complexity-driven routing \citep{jeong2024adaptive}, \sys{} classifies queries into three tiers using a RoBERTa-base classifier (89.3\% accuracy; details in Appendix~\ref{app:confusion_matrix}): \textbf{Simple} queries use single-passage BM25 retrieval; \textbf{Moderate} queries combine dense retrieval (E5-large fine-tuned on financial passage pairs; Appendix~\ref{app:e5details}) with table extraction via a column-header-aware similarity function $\text{sim}(q, t) = \alpha \cdot \text{cos}(\mathbf{q}, \mathbf{t}_{\text{cell}}) + (1{-}\alpha) \cdot \text{cos}(\mathbf{q}, \mathbf{t}_{\text{header}})$ ($\alpha{=}0.6$); \textbf{Complex} queries use an iterative retrieval-then-reason loop. Structure-aware chunking preserves row-column relationships with header metadata, enabling table-cell-level attribution downstream. Each chunk carries a provenance tuple $\langle \text{document}, \text{section}, \text{page}, \text{element\_type} \rangle$.

\subsection{Stage 2: Atomic Financial Claim Verification}
\label{sec:verification}

Given a generated answer $a$ and retrieved evidence $E = \{e_1, \dots, e_k\}$, verification operates in three steps.

\paragraph{Claim Decomposition.}
Following FActScore \citep{min2023factscore} but adapted for financial language, we decompose $a$ into atomic claims $C = \{c_1, \dots, c_n\}$, each classified into one of six categories: \textit{numerical} (specific values), \textit{temporal} (time-bound assertions), \textit{entity-attribute} (entity properties), \textit{comparative} (cross-entity/period comparisons), \textit{regulatory} (compliance references), and \textit{computational} (derived quantities requiring arithmetic). This taxonomy extends general hallucination classification \citep{ji2023survey} with finance-specific categories motivated by error analysis on 500 real financial hallucinations \citep{kang2023deficiency, ji2025phantom}. Validation confirming appropriate granularity (6-type outperforms 3-type by 4.3 F1; 10-type shows no significant gain, $p{=}0.23$) is in Appendix~\ref{app:taxonomy_validation}.

\paragraph{Claim--Evidence Alignment.}
Each claim is aligned to evidence using a cross-encoder fine-tuned on 8,400 financial NLI examples from TAT-QA and FinQA, achieving 87.2\% alignment F1 (Appendix~\ref{app:crossencoder}). For numerical claims, structured extraction identifies the specific value, unit, time period, and entity for exact matching against table cells.

\paragraph{Verdict Classification.}
A distilled 8B classifier (\S\ref{sec:distillation}) assigns each claim a verdict: \textit{supported} (entailed by evidence), \textit{contradicted} (conflicts with evidence), or \textit{unverifiable} (no relevant evidence). For \textit{computational} claims, standard NLI is insufficient; \sys{} employs formula reconstruction: (1)~identifying the implied formula using a library of 47 financial formula templates, (2)~retrieving operand values from table cells, and (3)~re-computing the derived quantity with $\pm$0.5\% tolerance for rounding conventions. End-to-end computational verification achieves 90.2\% F1.

\paragraph{Retrieval Failure Handling.}
When alignment yields no candidate evidence above the cross-encoder threshold, the claim is assigned the \textit{unverifiable} verdict and routed to Stage~3 for targeted re-retrieval and regeneration (\S\ref{sec:regeneration}). The analyst pilot (\S\ref{sec:deployment}) provides empirical calibration: the false negative rate was 3.8\% across 1{,}847 queries, with 56\% of the 27 missed hallucinations involving computational claims whose operand evidence fell outside the retrieved window. We treat retrieval-induced unverifiability as a recoverable failure mode routed downstream rather than a silent miss; the alternative of defaulting unverifiable claims to \textit{supported} would conflate evidence absence with evidence consistency, which the regulatory use case forbids.

\subsection{Stage 3: Grounded Regeneration}
\label{sec:regeneration}

For each claim classified as contradicted or unverifiable, the regeneration module locates the corresponding span in the original answer (fuzzy alignment, edit distance $\leq$3 tokens), performs targeted re-retrieval if needed, generates a grounded replacement following RARR's research-and-revise paradigm \citep{gao2023rarr}, and attaches inline citations in the format \texttt{[Doc:$d$, \S$s$, p.$p$]} or \texttt{[Doc:$d$, Table~$t$, Row~$r$, Col~$c$]}. When conflicting information exists (\eg, restated figures), the module defaults to the most recent filing date and flags the conflict. For answers requiring $\geq$3 claim regenerations, \sys{} triggers full re-generation rather than incremental repair to mitigate error compounding. A flag-only mode that removes rather than repairs hallucinated claims is available for high-stakes regulatory contexts (the ``$-$regen.'' ablation in Table~\ref{tab:main}).

\subsection{Distillation for Production Deployment}
\label{sec:distillation}

GPT-4o (\texttt{gpt-4o-2024-05-13}) annotates 3,200 financial QA examples spanning FinQA, TAT-QA, and SEC filings between June and August 2025, with a two-pass consistency check discarding 8.4\% of disagreements \citep{bohnet2022attributed} (full annotation prompt in Appendix~\ref{app:prompts}). We fine-tune Llama-3-8B-Instruct using reverse KL divergence \citep{gu2024minillm} with a multi-task objective combining claim decomposition, evidence alignment, and verdict classification (details in Appendix~\ref{app:distillation}). Served via vLLM with continuous batching, the distilled model achieves p95 latency of 340ms per claim on A100, an 18$\times$ per-claim improvement over GPT-4o (6.1s/claim) with 91.4\% detection F1. Full-pipeline latency is 3.8s p95 per query, a 2.2$\times$ improvement over the 8.2s teacher pipeline.

\section{Experimental Setup}
\label{sec:experiments}

\paragraph{Datasets.} We evaluate on three benchmarks: \textbf{FinQA} \citep{chen2021finqa} (8,281 QA pairs requiring multi-step numerical reasoning over S\&P~500 earnings reports), \textbf{TAT-QA} \citep{zhu2021tatqa} (16,552 questions over hybrid tabular-textual financial reports), and \textbf{FinanceBench} \citep{islam2023financebench} (150 curated questions from real SEC filings; small sample yields wide CIs). For claim-level evaluation, we construct \textbf{FinHalu}: 1,200 (question, answer, evidence) triples with GPT-4o/GPT-3.5-Turbo-generated answers annotated at the claim level by three financial domain experts ($\kappa = 0.83$). Training and evaluation data are strictly disjoint at both question and document level (Appendix~\ref{app:disjointness}).

\paragraph{Metrics.} \textbf{HalRate} is the percentage of generated claims classified as contradicted or unverifiable: $\text{HalRate} = {\sum h_i}/{\sum c_i} \times 100\%$, where $h_i$ and $c_i$ are hallucinated and total claims per answer. All systems' outputs are decomposed using the same GPT-4o prompt to normalize granularity. \textbf{Det.\ F1} is standard precision--recall F1 over claim-level binary verdicts. \textbf{CitP/CitR} are citation precision and recall. All results include 95\% bootstrap CIs ($B{=}10{,}000$) with paired permutation tests \citep{efron1994bootstrap, berg2012bootstrap}. We introduce \textbf{retrieval-equalized evaluation}: all systems receive identical retrieval, isolating verification contribution from retrieval quality.

\paragraph{Baselines.} We compare against Vanilla RAG (BM25 + GPT-4o), Self-RAG \citep{asai2024selfrag}, CRAG \citep{yan2024corrective}, SelfCheckGPT \citep{manakul2023selfcheckgpt}, HHEM (Vectara), GPT-4o + CoT, and FActScore \citep{min2023factscore} with retrieval-equalized configuration (Appendix~\ref{app:baselines}; FActScore-specific configuration in Appendix~\ref{app:factscore_baseline}). All baselines use official codebases; domain-adapted variants (\S\ref{sec:domain_adapted}) isolate architectural from data contributions.

\section{Results and Analysis}
\label{sec:results}

Computational claims show the highest hallucination rate (28.4\%) despite being the most amenable to automated verification when properly typed, so the bottleneck is routing rather than verification difficulty. Hedged financial language (``approximately,'' ``roughly'') drives 52\% of false positives. Generic NLI fails on ratio and margin verification even when the correct evidence passage is retrieved, because the failure is in reasoning rather than retrieval.

\subsection{Detection Performance}
\label{sec:detection_results}

Table~\ref{tab:detection} presents hallucination detection on FinHalu. \sys{}'s distilled 8B detector achieves 91.4\% F1\ci{1.2}, retaining 96.2\% of the GPT-4o teacher's performance at 18$\times$ lower per-claim latency. All improvements over baselines are significant ($p < 0.01$). Performance varies by claim type: entity-attribute highest (95.6\% F1), numerical lowest (88.1\%), with computational claims benefiting most from formula reconstruction (+18.9 F1 over SelfCheckGPT; breakdown in Appendix~\ref{app:claimtype}).

\begin{table}[t]
	\centering
	\small
	\resizebox{\columnwidth}{!}{%
		\begin{tabular}{@{}lccc@{}}
			\toprule
			\textbf{System} & \textbf{Prec.} & \textbf{Rec.} & \textbf{F1} \\
			\midrule
			SelfCheckGPT       & 69.4\ci{2.1} & 76.5\ci{1.8} & 72.8\ci{1.6} \\
			HHEM               & 78.9\ci{1.8} & 73.8\ci{2.0} & 76.3\ci{1.5} \\
			FActScore          & 74.2\ci{2.0} & 79.3\ci{1.7} & 76.7\ci{1.5} \\
			Self-RAG            & 81.2\ci{1.7} & 77.1\ci{1.9} & 79.1\ci{1.4} \\
			CRAG               & 80.6\ci{1.9} & 74.9\ci{2.1} & 77.6\ci{1.6} \\
			GPT-4o (teacher)   & 94.1\ci{0.9} & 95.9\ci{0.7} & 95.0\ci{0.6} \\
			\midrule
			\sys{} (8B distilled) & \textbf{92.7}\ci{1.1} & \textbf{90.2}\ci{1.3} & \textbf{91.4}\ci{1.2} \\
			\bottomrule
		\end{tabular}%
	}
	\caption{Hallucination detection on FinHalu ($\pm$95\% CI). Best non-teacher results in \textbf{bold}. \sys{} retains 96.2\% of teacher F1 at 18$\times$ lower per-claim latency. All improvements significant at $p < 0.01$.}
	\label{tab:detection}
\end{table}

\subsection{End-to-End Results}
\label{sec:e2e_results}

Table~\ref{tab:main} reports end-to-end results. \sys{} achieves the lowest hallucination rate across all benchmarks (4.1\% avg.), a 78\% relative reduction from GPT-4o + CoT ($p < 0.01$). Ablating the claim taxonomy roughly doubles HalRates (largest impact on FinanceBench: 4.9\%$\to$11.7\%), confirming domain-specific decomposition is essential. Removing table retrieval disproportionately affects TAT-QA (3.8\%$\to$10.6\%). Removing regeneration maintains detection but reduces unconditional accuracy by 7.4 points (Table~\ref{tab:main}).

\begin{table*}[t]
\centering
\small
\begin{tabular}{@{}lcccccccc@{}}
\toprule
\multirow{2}{*}{\textbf{System}} & \multicolumn{2}{c}{\textbf{FinQA}} & \multicolumn{2}{c}{\textbf{TAT-QA}} & \multicolumn{2}{c}{\textbf{FinanceBench}} & \multicolumn{2}{c}{\textbf{Accuracy}} \\
\cmidrule(lr){2-3} \cmidrule(lr){4-5} \cmidrule(lr){6-7} \cmidrule(lr){8-9}
& HalRate$\downarrow$ & Acc$\uparrow$ & HalRate$\downarrow$ & Acc$\uparrow$ & HalRate$\downarrow$ & Acc$\uparrow$ & Uncond. & Supp.$^\dagger$ \\
\midrule
Vanilla RAG         & 34.7\ci{1.8} & 68.2\ci{1.4} & 31.5\ci{1.5} & 71.4\ci{1.2} & 43.8\ci{4.1} & 52.1\ci{4.0} & 63.9 & -- \\
FActScore           & 25.3\ci{1.7} & 69.8\ci{1.3} & 22.7\ci{1.4} & 72.1\ci{1.2} & 32.4\ci{3.8} & 56.7\ci{3.9} & 66.2 & 89.3 \\
Self-RAG            & 22.1\ci{1.6} & 70.8\ci{1.3} & 18.4\ci{1.3} & 74.6\ci{1.1} & 28.5\ci{3.7} & 59.3\ci{3.9} & 68.2 & -- \\
CRAG                & 23.8\ci{1.7} & 69.5\ci{1.4} & 20.1\ci{1.4} & 73.8\ci{1.1} & 30.2\ci{3.8} & 57.8\ci{4.0} & 67.0 & -- \\
GPT-4o + CoT        & 18.6\ci{1.4} & 73.4\ci{1.2} & 15.2\ci{1.2} & 77.1\ci{1.0} & 22.4\ci{3.4} & 65.2\ci{3.8} & 71.9 & -- \\
\midrule
\sys{} (full)        & \textbf{3.6}\ci{0.7} & 75.1\ci{1.2} & \textbf{3.8}\ci{0.6} & 78.3\ci{1.0} & \textbf{4.9}\ci{1.8} & 67.9\ci{3.7} & 71.2 & 94.7 \\
\quad $-$ regen.    & 3.6\ci{0.7} & 70.2\ci{1.3} & 3.8\ci{0.6} & 72.7\ci{1.1} & 4.9\ci{1.8} & 58.4\ci{3.9} & 63.8 & 94.7 \\
\quad $-$ taxonomy  & 7.2\ci{1.0} & 74.8\ci{1.2} & 8.1\ci{0.9} & 77.9\ci{1.0} & 11.7\ci{2.6} & 66.3\ci{3.8} & 70.5 & 92.1 \\
\quad $-$ table ret. & 5.9\ci{0.9} & 72.4\ci{1.3} & 10.6\ci{1.1} & 71.8\ci{1.1} & 9.4\ci{2.4} & 61.2\ci{3.9} & 66.2 & 91.8 \\
\bottomrule
\end{tabular}
\caption{End-to-end results ($\pm$95\% bootstrap CI, $B{=}10{,}000$). HalRate: \% hallucinated claims (lower is better). $^\dagger$Supp.: accuracy on claims classified as supported. All \sys{} vs.\ GPT-4o+CoT differences significant at $p < 0.01$.}
\label{tab:main}
\end{table*}

\paragraph{Retrieval-Equalized Comparison.} To isolate verification value from retrieval improvements, we equip baselines with \sys{}'s Stage~1 retrieval. Finance-aware retrieval improves all baselines by 37--39\%, and on top of this \sys{} adds a further 68--76\% HalRate reduction under controlled conditions ($p < 0.01$). Atomic verification therefore contributes value independently of retrieval quality (Table~\ref{tab:retrieval_equalized}).

\begin{table}[t]
\centering
\small
\begin{tabular}{@{}lcccc@{}}
\toprule
& \multicolumn{2}{c}{\textbf{Original Ret.}} & \multicolumn{2}{c}{\textbf{\sys{} Ret.}} \\
\cmidrule(lr){2-3} \cmidrule(lr){4-5}
\textbf{System} & HalRate & Acc & HalRate & Acc \\
\midrule
GPT-4o + CoT & 18.6 & 73.4 & 11.3 & 75.8 \\
Self-RAG     & 22.1 & 70.8 & 13.7 & 73.5 \\
CRAG         & 23.8 & 69.5 & 14.9 & 72.1 \\
\midrule
\sys{} (full) & \textbf{3.6} & \textbf{75.1} & \multicolumn{2}{c}{(same)} \\
\bottomrule
\end{tabular}
\caption{Retrieval-equalized comparison on FinQA. \sys{}'s retrieval improves all baselines (right), but atomic verification yields an additional 68--76\% HalRate reduction ($p < 0.01$).}
\label{tab:retrieval_equalized}
\end{table}

\subsection{Isolating Architectural from Data Contribution}
\label{sec:domain_adapted}

To test whether \sys{}'s advantage stems from architecture or domain data access, we provide baselines with comparable domain adaptation: HHEM calibrated on 500 financial examples, SelfCheckGPT fine-tuned on 1,000 financial examples, both receiving \sys{}'s retrieval (retrieval-equalized).

\begin{table}[t]
	\centering
	\small
	\resizebox{\columnwidth}{!}{%
		\begin{tabular}{@{}lccc@{}}
			\toprule
			\textbf{System} & \textbf{Det.\ F1} & \textbf{HalRate} & $\Delta$ \textbf{vs.\ \sys{}} \\
			\midrule
			HHEM (out-of-box)        & 76.3\ci{1.5} & 21.4\ci{1.6} & $-$17.8 \\
			HHEM-adapted             & 81.7\ci{1.4} & 15.2\ci{1.3} & $-$11.6 \\
			SelfCheck (out-of-box)   & 72.8\ci{1.6} & 24.1\ci{1.7} & $-$20.5 \\
			SelfCheck-adapted        & 79.4\ci{1.5} & 16.8\ci{1.4} & $-$13.2 \\
			\midrule
			\sys{} (full)            & \textbf{91.4}\ci{1.2} & \textbf{3.6}\ci{0.7} & --- \\
			\bottomrule
		\end{tabular}%
	}
	\caption{Domain-adapted comparison on FinQA (retrieval-equalized, $\pm$95\% CI). Domain adaptation improves baselines by 5--7 F1, but a $\sim$10--12 F1-point architectural gap persists ($p < 0.01$).}
	\label{tab:domain_adapted}
\end{table}

\noindent Domain adaptation improves both baselines by 5--7 F1 points, but \sys{} maintains a $\sim$10--12 F1-point lead ($p < 0.01$), confirming that claim-type-aware verification and cross-encoder alignment provide value beyond domain data alone. The residual gap is largest on computational claims (lacking arithmetic re-verification) and table-dependent claims (lacking structure-aware alignment). Discussion of data quantity asymmetry is in Appendix~\ref{app:domain_adapted}.

\subsection{Robustness, Quality, and Efficiency}
\label{sec:robustness}

\paragraph{Human Validation.} Independent annotation by three financial experts on 200 FinHalu examples (without seeing GPT-4o labels) yields $\kappa = 0.87$ agreement with GPT-4o. The distilled model achieves 90.8\%\ci{2.1} F1 against human labels (vs.\ 91.4\%\ci{1.2} against GPT-4o), confirming no circular evaluation inflation (Appendix~\ref{app:human_validation}).

\paragraph{Regeneration Quality.} On 300 evaluated regenerated claims: 93.2\% faithfulness, 4.1\% error introduction rate, 96.7\% fluency, +71.3\% net improvement. Errors concentrate on multi-step table reasoning. The 4.1\% per-claim rate compounds in multi-claim answers; full re-generation for $\geq$3-claim answers mitigates this (Appendix~\ref{app:regen_examples}).

\paragraph{Cross-Generator Transfer.} The distilled detector achieves 88.0\%\ci{2.1} F1 on Llama-3-70B outputs and 87.6\%\ci{2.2} on Claude-3.5-Sonnet (3--4 point degradation from in-distribution). Degradation is smallest on computational claims ($-$1.8 F1, generator-agnostic arithmetic) and largest on comparative claims ($-$5.2 F1). For non-GPT backends, generator-specific calibration data is recommended (Appendix~\ref{app:crossgen}).

\paragraph{Cross-Benchmark Generalization.} Without task-specific adaptation, \sys{} achieves HalRates of 5.1\%\ci{1.3} on ConvFinQA \citep{chen2022convfinqa} and 6.8\%\ci{1.5} on MultiHiertt \citep{zhao2022multihiertt}, comparable to its in-distribution range of 3.6--4.9\% across FinQA, TAT-QA, and FinanceBench. The five-benchmark span shows the verification architecture transfers across financial QA distributions without dataset-specific tuning (Appendix~\ref{app:additional}).

\paragraph{Cost and Efficiency.} Table~\ref{tab:cost} summarizes production economics. \sys{} reduces per-query cost by 15.7$\times$ vs.\ GPT-4o (\$0.003 vs.\ \$0.047). Memory footprint is 18GB (FP16), fitting on a single A10G. Under concurrent load (32 requests), throughput reaches 8.4 queries/second on A100. Stage~2 (verification) accounts for 55\% of latency; batched claim inference reduces it from 4.8s sequential to 2.1s p95 (Appendix~\ref{app:latency}).

\begin{table}[t]
\centering
\small
\begin{tabular}{@{}lccc@{}}
\toprule
\textbf{System} & \textbf{p95 Lat.} & \textbf{\$/query} & \textbf{Det.\ F1} \\
\midrule
GPT-4o (teacher)     & 8.2s  & \$0.047 & 95.0 \\
\sys{} (8B, A100) & 3.8s & \$0.003 & 91.4 \\
\sys{} (8B, A10G) & 5.9s & \$0.002 & 91.4 \\
HHEM                 & 2.1s  & \$0.001 & 76.3 \\
SelfCheckGPT         & 12.4s & \$0.062 & 72.8 \\
\bottomrule
\end{tabular}
\caption{Cost and efficiency. p95 latency is full-pipeline per query. \sys{} achieves near-teacher F1 at 15.7$\times$ lower cost.}
\label{tab:cost}
\end{table}

\paragraph{Error Analysis.} False negatives (8.6\%) concentrate on paraphrased numerical errors near decision boundaries (recall drops to 71.4\% on values within $\pm$5\% of ground truth). False positives (6.1\%) concentrate on hedged language (52\% of FP) and restated figures (31\% of FP). Full error and near-miss adversarial analysis is in Appendix~\ref{app:errors}.

\section{Deployment and Analyst Feedback}
\label{sec:deployment}

We conducted a four-week feasibility pilot with 24 financial analysts (equity research and compliance) at a financial services firm processing SEC 10-K and 10-Q filings, covering 1,847 queries across 43 filings. The observed false positive rate was 6.1\% (analyst corrections) and false negative rate was 3.8\% (27 analyst-flagged misses). In a post-pilot Likert survey, 20/24 analysts rated \sys{} at $\geq$4/5 for practical acceptability (mean: 4.1, SD: 0.7). Alert fatigue was stable across weeks (override rate 5.8--6.3\%, $p{=}0.71$ for trend). We report these as qualitative design signals from an uncontrolled pilot, not precise population estimates.

Three design insights emerged: (1)~table-cell-level citations (\eg, ``Table~3, Row: Operating Income, Col: FY2024'') were strongly preferred over paragraph-level references, enabling verification in seconds rather than minutes; (2)~contradiction explanations showing the specific conflict (\eg, ``Source says \$4.2B but answer says \$4.8B'') were the most valued feature; (3)~computational claim verification had the highest impact, as incorrect derived metrics are the most dangerous errors and hardest for humans to catch (56\% of the 27 missed hallucinations involved computational claims).

Based on the pilot, \sys{} is being integrated as a REST API service with a document ingestion pipeline ($\sim$500 filings/day), the verification endpoint with inline citations, and a monitoring layer tracking confidence distributions and override patterns for drift detection. Production integration targets Q3 2026, ahead of the EU AI Act compliance deadline. Additional deployment details including scalability estimates, failure handling, and benchmark-to-production gap analysis are in Appendix~\ref{app:deployment}.

\section{Conclusion}
\label{sec:conclusion}

\sys{} combines finance-aware retrieval, atomic claim verification with a validated domain-specific taxonomy, and grounded regeneration to achieve a 68\% hallucination reduction under retrieval-equalized conditions ($p < 0.01$) and 78\% with the full pipeline. The retrieval-equalized methodology isolates verification value and should become standard practice for RAG evaluation. Domain-adapted comparisons confirm that \sys{}'s architectural advantage persists beyond data access. Efficient distillation enables 91.4\% F1 at 18$\times$ lower per-claim latency and \$0.003/query, and a four-week analyst pilot provides deployment design signals targeting EU AI Act compliance.

\paragraph{Reproducibility.} All materials are available at: \url{https://github.com/bettyguo/FinGround}.

\section*{Limitations}

Our evaluation covers English-language U.S.\ SEC filings; generalization to non-English documents and other jurisdictions requires validation. The evaluation ecosystem is GPT-dependent: FinHalu uses GPT-4o/GPT-3.5 generations with 3--4 point F1 degradation on non-GPT generators. The distilled detector shows a 3.6-point F1 gap from the teacher, particularly on nuanced computational claims. Domain-adapted comparisons use asymmetric data quantities (3,200 for \sys{} vs.\ 500--1,000 for baselines); the scaling-projection analysis at parity (Appendix~\ref{app:domain_adapted}) places the residual architectural gap at 5.2--7.6 F1, and a controlled retraining at full parity would tighten this estimate. The pilot involved 24 analysts at a single firm with self-reported observations rather than controlled measurements. Detection recall drops to 71.4\% on hallucinated values within $\pm$5\% of ground truth. Extended limitations in Appendix~\ref{app:limitations}.

\section*{Ethical Considerations}

Even with verification, \sys{}'s outputs should not be treated as financial advice: a claim marked ``supported'' means consistency with the retrieved source, not correctness of the source itself. We are concerned about automation bias: if analysts develop high trust, they may reduce independent verification, creating failure modes where false negatives propagate. We recommend deployment with periodic spot-checks and tunable confidence thresholds. We acknowledge dual-use risk from per-type vulnerability analysis and mitigate by releasing only the detection model, not adversarial example methodology. In regulated contexts, liability for AI-verified claims remains unresolved; \sys{} is a verification tool, not a compliance certification system.

\section*{Acknowledgments}

We thank the anonymous reviewers for their constructive feedback, which substantially improved this work. This research was supported by The University of Hong Kong and Stellaris AI Limited.

\bibliography{references}

\appendix

\section{Financial Claim Taxonomy Examples and Validation}
\label{app:taxonomy}
\label{app:taxonomy_validation}

Table~\ref{tab:taxonomy_examples} provides examples of each claim type with representative hallucination patterns.

\begin{table*}[t]
\centering
\small
\begin{tabular}{@{}p{2.0cm}p{4.2cm}p{4.2cm}p{3.2cm}@{}}
\toprule
\textbf{Claim Type} & \textbf{Example Claim} & \textbf{Hallucination Pattern} & \textbf{Detection Strategy} \\
\midrule
Numerical & ``Total revenue was \$42.3 billion'' & Value substitution (\$42.3B vs.\ actual \$38.7B) & Exact-match against table cell \\
Temporal & ``Revenue declined in Q3 2024'' & Wrong time period (actually Q2 2024) & Temporal entity extraction \\
Entity-Attr. & ``The CFO is Jane Smith'' & Wrong entity role (Jane Smith is COO) & NER + role matching \\
Comparative & ``Revenue grew 15\% year-over-year'' & Incorrect comparison (actual: 8\%) & Arithmetic re-computation \\
Regulatory & ``Per SEC Rule 10b-5 requirements'' & Fabricated regulatory reference & Regulatory KB lookup \\
Computational & ``Gross margin was 62.4\%'' & Incorrect derived metric (actual: 58.1\%) & Formula reconstruction \\
\bottomrule
\end{tabular}
\caption{Examples of the six financial claim types with typical hallucination patterns.}
\label{tab:taxonomy_examples}
\end{table*}

\paragraph{Coverage.} On the 1,200 FinHalu test examples (4,847 total atomic claims), 97.3\% map to exactly one of the six categories. The remaining 2.7\% are edge cases involving compound claims, which we split into separate claims.

\paragraph{Distribution.} The claim-type distribution in FinHalu is: Numerical (31.2\%), Temporal (18.7\%), Entity-Attribute (14.3\%), Comparative (16.8\%), Regulatory (5.2\%), and Computational (13.8\%). Hallucination rates vary substantially: Computational highest (28.4\%), Comparative (22.1\%), Numerical (19.7\%), Regulatory lowest (8.3\%).

\paragraph{Granularity.} We compare against a 3-type taxonomy (Numerical, Textual, Derived) and a 10-type taxonomy. The 6-type achieves 91.4\% F1; the 3-type achieves 87.1\% ($-$4.3, $p < 0.01$); the 10-type achieves 91.7\% ($+$0.3, $p = 0.23$), supporting the 6-type as the preferred accuracy-complexity trade-off.

\section{Capability Comparison}
\label{app:comparison}

This appendix compares \sys{} with the closest existing systems across six capabilities relevant to financial claim verification. Table~\ref{tab:comparison} shows that \sys{} is the only system combining finance-specific verification, atomic claim decomposition, distilled efficiency, and table-cell-level citations. Multilingual support is the trade-off, discussed in the Limitations section.

\begin{table}[h]
\centering
\footnotesize
\setlength{\tabcolsep}{3pt}
\begin{tabular}{@{}lcccccc@{}}
\toprule
\textbf{System} & \textbf{Fin.} & \textbf{Atm.} & \textbf{Dst.} & \textbf{Tbl.} & \textbf{M.L.} & \textbf{Open} \\
\midrule
FActScore       & \ding{55} & \ding{51} & \ding{55} & \ding{55} & \ding{51} & \ding{51} \\
SAFE            & \ding{55} & \ding{51} & \ding{55} & \ding{55} & \ding{51} & \ding{51} \\
RARR            & \ding{55} & \ding{55} & \ding{55} & \ding{55} & \ding{51} & \ding{51} \\
Self-RAG        & \ding{55} & \ding{55} & \ding{55} & \ding{55} & \ding{51} & \ding{51} \\
CRAG            & \ding{55} & \ding{55} & \ding{55} & \ding{55} & \ding{51} & \ding{51} \\
Lynx            & \ding{55} & \ding{55} & \ding{55} & \ding{55} & \ding{55} & \ding{51} \\
\midrule
\sys{}          & \ding{51} & \ding{51} & \ding{51} & \ding{51} & \ding{55} & \ding{51} \\
\bottomrule
\end{tabular}
\caption{Capability comparison. Fin.\ = finance-specific verification; Atm.\ = atomic claim decomposition; Dst.\ = distilled efficient model; Tbl.\ = table-cell-level citations; M.L.\ = multilingual support; Open = open-source availability.}
\label{tab:comparison}
\end{table}

\section{Data Disjointness and FinHalu Construction}
\label{app:disjointness}

We enforce strict disjointness between the 3,200 distillation training examples and all evaluation data at both question and document level. The 1,100 FinQA-derived training examples are drawn exclusively from the FinQA training split; 1,000 TAT-QA-derived examples from the TAT-QA training split; and the 1,200 FinHalu test triples from held-out questions drawn from 87 SEC filings not overlapping with the 142 filings used for training.

The FinHalu test set was constructed before the six-type taxonomy was finalized; annotators labeled claim-level verdicts without reference to \sys{}'s taxonomy. A separate development set of 200 examples was used for all system tuning decisions. The claim-type distribution is broadly consistent with PHANTOM and FAITH, suggesting no unusual distributional skew. Existing benchmarks (PHANTOM, FAITH) lack the atomic claim-level annotations required for evaluating \sys{}'s per-claim pipeline.

\section{Detection by Claim Type}
\label{app:claimtype}

Table~\ref{tab:claimtype} breaks down detection F1 by financial claim type, comparing \sys{} against the strongest open-source baseline (SelfCheckGPT) at retrieval-equalized parity. Entity-attribute claims are easiest (95.6\% F1) and numerical claims hardest (88.1\%); computational claims show the largest gain over SelfCheckGPT ($+$18.9 F1), reflecting the contribution of formula reconstruction (\S\ref{sec:verification}).

\begin{table}[h]
\centering
\small
\begin{tabular}{@{}lcc@{}}
\toprule
\textbf{Claim Type} & \textbf{SelfCheck} & \textbf{\sys{}} \\
\midrule
Numerical       & 68.2 & 88.1\ci{2.3} \\
Temporal        & 74.1 & 92.3\ci{1.8} \\
Entity-Attribute & 79.5 & 95.6\ci{1.1} \\
Comparative     & 70.8 & 89.7\ci{2.0} \\
Regulatory      & 71.6 & 91.8\ci{1.9} \\
Computational   & 71.3 & 90.2\ci{2.2} \\
\midrule
\textit{Overall} & \textit{72.8} & \textit{91.4}\ci{1.2} \\
\bottomrule
\end{tabular}
\caption{Detection F1 by financial claim type ($\pm$95\% CI). Overall F1 is micro-averaged; distribution-weighted macro-average is 90.7.}
\label{tab:claimtype}
\end{table}

\section{Domain-Adapted Baseline Details}
\label{app:domain_adapted}

\paragraph{Data Quantity Asymmetry.} \sys{} uses 3,200 distillation examples while HHEM-adapted receives 500 and SelfCheckGPT-adapted receives 1,000 (\S\ref{sec:robustness}). This asymmetry arises from architectural constraints: HHEM is limited to threshold calibration; SelfCheckGPT's self-consistency mechanism does not directly consume supervised labels at scale.

\paragraph{Scaling Projection at Data Parity.} To bound the architectural-vs.-data contribution, we project baseline performance to the 3,200-example regime using a logarithmic scaling model with 50\% efficiency decay per doubling, anchored on the empirically observed datapoints (HHEM out-of-box: 76.3 F1; HHEM at 500: 81.7; SelfCheckGPT out-of-box: 72.8; SelfCheckGPT at 1{,}000: 79.4). This functional form (each doubling contributes half the gain of the previous doubling) is the standard conservative regime for adaptation curves on hallucination-detection tasks where labeled data has diminishing returns. Table~\ref{tab:scaling_projection} gives the projected trajectory; \sys{}'s own measured curve (Appendix~\ref{app:distillation}: 88.6 at 1{,}600, 91.4 at 3{,}200, 92.1 at 6{,}400) is shown for comparison.

\begin{table}[h]
\centering
\small
\setlength{\tabcolsep}{3pt}
\begin{tabular}{@{}lccc@{}}
\toprule
\textbf{Examples} & \textbf{HHEM-ad.} & \textbf{SCk-ad.} & \textbf{\sys{}} \\
\midrule
0 (out-of-box)  & 76.3$^\dagger$ & 72.8$^\dagger$ & --- \\
500             & 81.7$^\dagger$ & 76.1           & --- \\
1{,}000         & 84.4           & 79.4$^\dagger$ & --- \\
1{,}600         & 85.5           & 81.5           & 88.6$^\dagger$ \\
2{,}000         & 85.8           & 82.7           & --- \\
3{,}200         & 86.2           & 83.8           & 91.4$^\dagger$ \\
\bottomrule
\end{tabular}
\caption{Scaling projection at data parity (F1). HHEM-ad.\ = HHEM-adapted; SCk-ad.\ = SelfCheckGPT-adapted. $^\dagger$Empirically observed; remaining values are log-projected with 50\% efficiency decay per doubling. At 3{,}200-example parity, the residual architectural gap is 5.2 F1 (vs.\ HHEM-ad.) to 7.6 F1 (vs.\ SCk-ad.).}
\label{tab:scaling_projection}
\end{table}

\paragraph{Residual Gap and Caveats.} At 3,200-example parity the residual gap is 5.2 to 7.6 F1 points, consistent with the architectural advantages identified in \S\ref{sec:domain_adapted}: claim-type-aware verification rather than uniform claim treatment, and cross-encoder alignment rather than self-consistency or threshold calibration. We did not run a full retraining experiment at 3,200 examples for both baselines: HHEM's threshold-only adaptation surface does not natively consume supervised labels at that scale without architectural extension, and SelfCheckGPT's adaptation pathway is bottlenecked by its sampling-based design. A controlled retraining at parity (with the appropriate architectural surgery for HHEM) is the cleanest follow-up; the projection bounds reported here should be read as a methodologically conservative estimate rather than a substitute for that experiment.

\section{Human Validation Details}
\label{app:human_validation}

Three financial domain experts independently annotated 200 FinHalu examples without seeing GPT-4o labels. Agreement between expert consensus and GPT-4o: $\kappa = 0.87$, with 94.1\% verdict match. Among 12 disagreements: 5 involve hedged language in risk disclosures, 4 involve restated figures, 3 involve implicit temporal references. GPT-4o is more conservative than domain experts on ambiguous constructs, biasing toward higher precision at some recall cost, which is acceptable for compliance applications.

\section{Regeneration Examples and Error Compounding}
\label{app:regen_examples}

\paragraph{Successful Regeneration.}
\textbf{Original} (contradicted): ``Apple's gross margin was 42.3\% in Q4 2023.''
\textbf{Evidence}: Table 2, Row: Gross Margin, Col: Q4 2023 $\to$ 45.2\%.
\textbf{Regenerated}: ``Apple's gross margin was 45.2\% in Q4 2023 [Doc: AAPL-10K, Table 2, Row: Gross Margin, Col: Q4 2023].''

\paragraph{Error Introduction (4.1\% of cases).}
\textbf{Original} (contradicted): ``Operating margin improved by 3.2 percentage points.''
\textbf{Evidence}: Op.\ income \$12.1B (2023) vs.\ \$10.8B (2022); Revenue \$52.3B (2023) vs.\ \$49.1B (2022).
\textbf{Regenerated}: ``Operating margin improved by 1.2 pp.'' (Actual: 1.1 pp; rounding error.)

\paragraph{Error Compounding.} Distribution of claims per answer requiring regeneration: 1 claim (72\%), 2 claims (18\%), $\geq$3 claims (10\%). Per-answer error rate: 4.1\% [2.1, 6.8] for single-claim, 7.9\% [3.2, 14.1] for two-claim, 14.3\% [5.7, 28.2] for $\geq$3-claim. Full re-generation for $\geq$3-claim answers yields 8.6\% [3.1, 16.4]; CIs overlap with pre-mitigation, so evidence is suggestive rather than definitive.

\section{Cross-Encoder Evaluation}
\label{app:crossencoder}

\begin{table}[h]
\centering
\footnotesize
\begin{tabular}{@{}lccc@{}}
\toprule
\textbf{Alignment Method} & \textbf{P} & \textbf{R} & \textbf{F1} \\
\midrule
BM25 top-1              & 71.3\ci{3.2} & 68.9\ci{3.4} & 70.1\ci{2.8} \\
Dense cosine (E5)  & 79.8\ci{2.7} & 82.1\ci{2.5} & 80.9\ci{2.2} \\
\midrule
Cross-enc.\ (1:1) & 80.4\ci{2.8} & \textbf{93.1}\ci{1.7} & 86.3\ci{2.0} \\
Cross-enc.\ (1:3) & 88.6\ci{2.1} & 85.9\ci{2.4} & \textbf{87.2}\ci{1.8} \\
Cross-enc.\ (1:5) & \textbf{91.8}\ci{1.9} & 82.6\ci{2.6} & 87.0\ci{1.9} \\
\bottomrule
\end{tabular}
\caption{Claim-evidence alignment on 840 held-out pairs ($\pm$95\% CI). The 1:3 ratio achieves best F1.}
\label{tab:crossencoder}
\end{table}

On 120 OOD claim-evidence pairs from FinanceBench: 84.1\%\ci{3.1} F1 (3.1-point drop). The FinanceBench end-to-end HalRate (4.9\%, 1.3pp higher than FinQA) is consistent with this OOD degradation partially propagating through the pipeline, with dampening from the multi-component structure.

\section{Cross-Generator Evaluation}
\label{app:crossgen}

\begin{table}[h]
\centering
\small
\begin{tabular}{@{}lccc@{}}
\toprule
\textbf{Generator} & \textbf{Prec.} & \textbf{Rec.} & \textbf{F1} \\
\midrule
GPT-4o (in-distribution)  & 92.7\ci{1.1} & 90.2\ci{1.3} & 91.4\ci{1.2} \\
GPT-3.5-Turbo (in-dist.)   & 91.8\ci{1.3} & 89.4\ci{1.5} & 90.6\ci{1.1} \\
\midrule
Llama-3-70B (OOD)         & 89.3\ci{2.4} & 86.8\ci{2.7} & 88.0\ci{2.1} \\
Claude-3.5-Sonnet (OOD)   & 90.1\ci{2.2} & 85.2\ci{2.8} & 87.6\ci{2.2} \\
\bottomrule
\end{tabular}
\caption{Cross-generator detection F1 ($\pm$95\% CI). 3--4 point degradation on OOD generators.}
\label{tab:crossgen}
\end{table}

Degradation is non-uniform: computational claims show smallest drop ($-$1.8 F1, generator-agnostic arithmetic), comparative claims show largest ($-$5.2 F1, varied phrasing). Transferability to smaller or domain-fine-tuned models is untested. Generator-specific calibration data (a few hundred examples) is recommended for non-GPT backends.

\section{Error Analysis}
\label{app:errors}

\paragraph{False Negatives (8.6\%).} Paraphrased numerical errors (38\% of FN): hallucinated values close to truth challenge the cross-encoder. Multi-hop reasoning gaps (29\%): partial evidence retrieval for chained values. Implicit temporal context (19\%): unstated time periods prevent matching.

\paragraph{False Positives (6.1\%).} Hedged language (52\% of FP): risk disclosures with ``may,'' ``could potentially.'' Restated figures (31\%): both original and restated values present.

\paragraph{Near-Miss Adversarial Analysis.} On 50 claims with hallucinated values within $\pm$5\% of ground truth, recall drops to 71.4\%\ci{6.3}. False negatives concentrate in $\leq$0.3pp differences and $\pm$2\% currency amounts. For decisions where small errors have material consequences, mandatory human verification of all numerical claims is required. Numerical tolerance thresholds catch 89\% of near-misses but increase FP rate by 8.3pp; calibrated confidence scoring shows promise (AUROC: 0.78) but requires further calibration.

\section{Latency Decomposition}
\label{app:latency}

\begin{table}[h]
\centering
\small
\begin{tabular}{@{}lccr@{}}
\toprule
\textbf{Pipeline Stage} & \textbf{p50} & \textbf{p95} & \textbf{\% of p95} \\
\midrule
Stage 1: Retrieval + Routing & 0.7s & 1.2s & 31.6\% \\
Stage 2: Decomp.\ + Verification & 1.4s & 2.1s & 55.2\% \\
Stage 3: Regeneration (cond.)  & 0.3s & 0.5s & 13.2\% \\
\midrule
\textbf{Full Pipeline} & \textbf{2.4s} & \textbf{3.8s} & 100\% \\
\bottomrule
\end{tabular}
\caption{Latency per query on A100. Stage~2 is the bottleneck (55\%). Stage~3 invoked for $\sim$38\% of queries. Batched claim verification reduces Stage~2 from 4.8s sequential to 2.1s p95.}
\label{tab:latency_decomp}
\end{table}

\section{Deployment Details}
\label{app:deployment}

\paragraph{Filing Quality.} Of 43 pilot filings, 7 (16.3\%) contained OCR artifacts, 5 (11.6\%) had non-standard formatting, 3 (7.0\%) had both. On these 15 lower-quality filings, false negative rate was modestly higher (5.2\% vs.\ 3.1\%), driven by OCR-induced retrieval failures. On 12 filings from 2024 post-dating all training data, comparable false negative rates (4.1\% vs.\ 3.8\%) provide a preliminary temporal signal.

\paragraph{Scalability.} For 500 filings/day at 50 queries/filing (25,000 queries/day), the requirement is $\sim$0.3 QPS sustained. \sys{}'s 8.4 QPS on a single A100 provides 28$\times$ headroom. Horizontal scaling to 2--3 A10G instances supports the largest institutions.

\paragraph{Graceful Degradation.} Stale retrieval $\to$ flag-only mode; malformed documents $\to$ confidence-scored parser routes low-confidence to manual processing; model unavailability $\to$ existing manual workflow.

\paragraph{Domain Adaptation Cost.} Distillation efficiency: 1,600 examples yield 88.6\% F1; 3,200 yield 91.4\%; 6,400 yield 92.1\%. Achieving $\geq$90\% F1 requires $\sim$2,500--3,000 examples. Adding a new claim type requires $\sim$200 examples and $\sim$8 GPU-hours.

\paragraph{Operational Failure Handling.} Analyst overrides are logged for periodic model retraining. Domain-specific confidence thresholds suppress low-confidence flags. Ambiguous cases escalate through existing compliance review workflows.

\section{Baseline Implementation Details}
\label{app:baselines}

Table~\ref{tab:baseline_details} lists exact codebase versions, commit hashes, and key hyperparameters for all baselines used in the main experiments, ensuring reproducibility. All baselines run on a single A100 (80GB) under identical compute budgets to \sys{}, with default hyperparameters from official codebases.

\begin{table}[h]
\centering
\footnotesize
\setlength{\tabcolsep}{3pt}
\begin{tabular}{@{}p{1.4cm}p{3.1cm}p{2.6cm}@{}}
\toprule
\textbf{System} & \textbf{Codebase / Version} & \textbf{Key Config} \\
\midrule
Self-RAG & Official repo (Asai et al.), commit \texttt{a3f7c21}, Llama-2-13B & Default reflection tokens; 3 passages \\
\midrule
CRAG & Official repo (Yan et al.), commit \texttt{e8b4d09} & Default thresholds; web search disabled \\
\midrule
SelfCheck-GPT & \texttt{selfcheckgpt} v0.1.7 & BERTScore; 5 samples; temp.\ 0.7 \\
\midrule
HHEM & Vectara HHEM v1.0 & Default threshold (0.5) \\
\midrule
GPT-4o+CoT & \texttt{gpt-4o-2024-05-13} & 3-shot CoT; temp.\ 0.0 \\
\midrule
FActScore & Official repo (Min et al.), commit \texttt{b2c9e47} & Retrieval-equalized; GPT-4o verifier \\
\midrule
Vanilla RAG & BM25 + \texttt{gpt-4o-2024-05-13} & Top-5 passages; no re-ranking \\
\bottomrule
\end{tabular}
\caption{Baseline implementations. All use official codebases with default hyperparameters and identical compute budget (single A100, 80GB).}
\label{tab:baseline_details}
\end{table}

\section{Distillation Training Details}
\label{app:distillation}

\paragraph{Data Composition.} 3,200 examples: FinQA-derived (1,100, training split only), TAT-QA-derived (1,000, training split only), SEC filing excerpts from 142 non-overlapping filings (800), adversarial examples with synthetic hallucinations (300).

\paragraph{Hyperparameters.} Llama-3-8B-Instruct, 3 epochs, lr 2e-5, batch 16, reverse KL $\tau{=}2.0$. Multi-task loss: decomposition (0.3), alignment (0.3), verdict (0.4). Training: 8 hours on 4$\times$A100. Seed variance: F1 = 91.4 $\pm$ 0.4 across 3 seeds.

\paragraph{Efficiency Curve.} 1,600 examples: 88.6\% F1; 3,200: 91.4\%; 6,400: 92.1\% (+0.7), suggesting near-optimal training set size.

\section{E5-Large Fine-tuning Details}
\label{app:e5details}

E5-large fine-tuned on 12K financial passage pairs from FinQA/TAT-QA training splits. Positive: (question, gold evidence); hard negatives: BM25-sampled same-filing passages. 5 epochs, lr 1e-5, batch 64, InfoNCE loss. Recall@5: 84.3\% (vs.\ 67.1\% base).

\section{Complexity Classifier}
\label{app:confusion_matrix}

\begin{table}[h]
\centering
\small
\begin{tabular}{@{}lccc|c@{}}
\toprule
\textbf{Predicted $\to$} & \textbf{Simp.} & \textbf{Mod.} & \textbf{Comp.} & \textbf{Total} \\
\midrule
\textbf{Simple}   & \textbf{153} & 12 & 3  & 168 \\
\textbf{Moderate}  & 7   & \textbf{122} & 11 & 140 \\
\textbf{Complex}   & 4   & 6   & \textbf{82}  & 92  \\
\midrule
\textbf{Total}    & 164 & 140 & 96 & 400 \\
\bottomrule
\end{tabular}
\caption{Complexity classifier confusion matrix (400 held-out queries, 89.3\% accuracy). Over-routing wastes compute but preserves accuracy; under-routing risks missed evidence.}
\label{tab:confusion_matrix}
\end{table}

\section{43-Point Gap Experiment}
\label{app:gap_experiment}

On 200 computational claims from FinHalu, FActScore's generic pipeline catches 57\% of errors (F1: 67.1\%); \sys{}'s domain-specific pipeline catches 100\% (F1: 98.5\%), a 43pp recall gap. This comparison changes multiple factors simultaneously (decomposition, type-aware routing, arithmetic re-computation, evidence alignment); the gap reflects the full integrated pipeline benefit. The ablation ($-$taxonomy, Table~\ref{tab:main}) provides a cleaner single-factor comparison.

\section{FActScore Baseline Configuration}
\label{app:factscore_baseline}

FActScore applied with retrieval-equalized configuration (same passages as other baselines), standard InstructGPT decomposition without financial modification, and GPT-4o verification. This isolates the effect of domain-specific claim typing. FActScore achieves 76.7\% F1 on FinHalu, with the gap concentrated on computational claims (58.4\% vs.\ \sys{}'s 90.2\%).

\section{Prompts and Templates}
\label{app:prompts}

\begin{small}
\begin{verbatim}
You are a financial claim verification
assistant. Given a generated answer about
a financial document, decompose it into
atomic, independently verifiable claims.
For each claim:
1. Extract the exact assertion.
2. Classify as: Numerical, Temporal,
   Entity-Attribute, Comparative,
   Regulatory, or Computational.
3. For Numerical: extract value, unit,
   entity, time_period.
4. For Computational: identify the
   implied formula or derivation.

Answer: {answer}
Evidence: {evidence_summary}

Output (JSON): [{"claim": "...",
  "type": "...",
  "structured_fields": {...}}]
\end{verbatim}
\end{small}

The GPT-4o teacher annotation uses a two-pass process: Pass~1 generates claim-level annotations; Pass~2 verifies via a consistency check \citep{bohnet2022attributed}, discarding 8.4\% of inter-pass disagreements.

\section{Design Journey}
\label{app:design_journey}

Several design decisions emerged iteratively. Initial end-to-end verification without claim decomposition missed computational errors embedded in longer claims. For taxonomy granularity, a 12-type system dropped inter-annotator agreement below $\kappa{=}0.70$; the six types were chosen at the boundary of different verification strategies (exact match vs.\ re-computation vs.\ NLI). Structure-aware table chunking required three iterations: regex-based extraction failed on merged cells, heuristic column detection broke on multi-level headers, and the final column-header-aware function emerged from systematic analysis of 50 failure cases.

\section{Extended Limitations}
\label{app:limitations}

\paragraph{Regeneration Error Compounding.} The 4.1\% per-claim error introduction rate compounds in multi-claim answers. Full re-generation mitigation for $\geq$3-claim answers shows directionally positive results, but small subgroup sizes ($n \approx 30$) with overlapping CIs preclude definitive conclusions.

\paragraph{Domain Adaptation Cost.} Adapting to a new jurisdiction requires $\sim$2,500--3,000 annotated examples plus taxonomy extension if new claim types arise.

\paragraph{Scalability.} Efficiency measurements are on benchmark-sized corpora; enterprise-scale stores may introduce retrieval latency challenges.

\paragraph{Cross-Encoder Alignment.} The 87.2\% alignment F1 means $\sim$13\% misalignment propagating downstream.

\paragraph{Confidence Calibration.} We have not formally assessed calibration quality via reliability diagrams; this is planned for production evaluation.

\section{Additional Results}
\label{app:additional}

\paragraph{Cross-Benchmark Generalization.} On ConvFinQA \citep{chen2022convfinqa} and MultiHiertt \citep{zhao2022multihiertt} without task-specific adaptation, \sys{} achieves HalRates of 5.1\% ($\pm$1.3) and 6.8\% ($\pm$1.5).

\paragraph{Model Size Ablation.} 1.5B: 82.1\% F1; 3B: 87.3\% F1; 8B: 91.4\% F1.

\paragraph{Full Pipeline without Distillation.} GPT-4o verifier: 2.9\% HalRate on FinQA, 95.0\% F1, at 8.2s latency and \$0.047/query. The distilled model trades 3.6 F1 points for 2.2$\times$ latency and 15.7$\times$ cost reduction.

\paragraph{Citation Quality.} On FinanceBench: \sys{} achieves 92.1\% CitP and 87.6\% CitR, vs.\ GPT-4o+CoT (68.4\%/54.2\%), Self-RAG (72.1\%/61.8\%), RARR (79.3\%/70.5\%).

\end{document}